\documentclass{article}

\usepackage[preprint]{neurips_2026}

\usepackage[utf8]{inputenc}
\usepackage[T1]{fontenc}
\usepackage{amsmath,amssymb,amsfonts}
\usepackage{bm}
\usepackage{booktabs}
\usepackage{hyperref}
\usepackage{url}
\usepackage{graphicx}
\usepackage{wrapfig}
\usepackage{enumitem}
\usepackage{silence}
\usepackage{microtype}
\usepackage{xcolor}
\usepackage{tikz}
\usetikzlibrary{positioning, arrows.meta, fit, backgrounds, calc}
\usepackage{algorithm}
\usepackage{algpseudocode}
\usepackage{float}
\usepackage{natbib}

\usepackage{xcolor}

\graphicspath{{plots/}}

\title{Reinforcement Learning with Decomposed Subtasks}

\author{%
  Mattie Terzolo \\
  Upwork \\
  \texttt{mattieterzolo@upwork.com} \\
  \And
  Mikolaj Sacha \\
  Upwork \\
  \texttt{mikolajsacha@cloud.upwork.com} \\
  \And
  Ayan Sinha \\
  Upwork \\
  \texttt{ayansinha@upwork.com} \\
  \And
  Andrew Rabinovich \\
  Upwork \\
  \texttt{andrewrabinovich@upwork.com} \\
}

\begin{document}

\maketitle

\begin{abstract}
Group Relative Policy Optimization (GRPO) and other reinforcement learning policy-gradient methods for training language model agents collapse an entire multi-turn rollout into a single scalar trajectory reward before that signal enters the policy update. When the underlying task is the composition of distinct skills, and especially when environmental feedback is sparse and delayed, this collapsing is lossy: the optimizer is left to implicitly infer which competency drove the outcome and how that should translate into behavioral change. We argue that the right primitive for these settings is not a better scalar but a decomposition: trajectory reward should be split along subtasks before it ever enters the policy update. We introduce \textbf{Reinforcement Learning with Decomposed Subtasks (RLDS)}, a recipe whose core is \textbf{Subtask-Decomposed Advantage Estimation (SDAE)}: a replacement for the scalar GRPO advantage that decomposes trajectory reward into per-subtask shares on a fixed subtask taxonomy, computes a group-relative advantage per subtask, and distributes per-token credit by weighting each subtask's advantage by its importance and concentrating it around the step where the reflection identifies that subtask's execution as being consequential to the outcome. We evaluate the resulting recipe on four agentic benchmarks spanning a range of complexity: FrozenLake (sparse grid navigation), HotpotQA (multi-hop QA with a single retrieval tool), ScienceWorld (long-horizon embodied science), and DeepResearch (long-form research with four tools and a composite rubric reward). Per-task heterogeneity diagnostics emitted during training show where decomposition pays off, and the empirical results match: gains scale with subtask heterogeneity, largest on the high-heterogeneity tasks ScienceWorld ($+11.5$ points, paired-bootstrap 95\% CI $[+9.8, +13.3]$) and FrozenLake ($+9.8$ points, $[+7.0, +12.8]$), and within noise on the lower-heterogeneity HotpotQA and DeepResearch, which the diagnostics flagged as having little for decomposition to recover. ScienceWorld is additionally more compute-efficient under RLDS than under scalar GRPO ($-10.9\%$ wall-clock per training step), as long rollouts amortize the fixed reflect-and-grade overhead.
\end{abstract}

\section{Introduction}
\label{sec:intro}

Consider an agent operating in ScienceWorld \citep{wang2022scienceworld}, where the action space is deep and rollouts compose multiple competencies. Two rollouts in the same group both end with reward 0.4. One fails to gather all the items needed to run the experiment; the other gathers the correct items but misinterprets the results. Group Relative Policy Optimization (GRPO) \citep{shao2024deepseekmath} subtracts the group mean, hands both rollouts the same advantage, and broadcasts that scalar uniformly across every token -- so the second rollout's correctly-gathered tokens get punished for its unrelated interpretation failure. A grader watching either rollout can name the failure in one sentence; the optimizer treats the two trajectories as equivalent. This is the central pathology of scalar reward in compositional RL: a single number per trajectory cannot route credit to the competency that needs it, and a uniform per-token broadcast cannot route credit to the segment that earned it. The motivation for treating this as a structural problem rather than a tuning problem comes from a position paper \citep{rabinovich2026rlwm}, which argues that scalar reward is an informational bottleneck on multi-dimensional tasks and that a task's dimensional structure must be made explicit within the training process itself.

This work addresses both facets of the pathology by giving the optimizer a structured diagnosis in place of a scalar. Reflection-augmented training methods (\S\ref{sec:bg}) already have the model grade its own rollout in natural language, but compress that grade into a flat trajectory reward before it reaches the optimizer. \textbf{Reinforcement Learning with Decomposed Subtasks (RLDS)} keeps the diagnosis structured end-to-end. The reflection phase is asked to grade each rollout along a fixed per-task taxonomy of competencies and to name, for each competency, the pivot step at which its outcome was determined; \textbf{Subtask-Decomposed Advantage Estimation (SDAE)} then routes the gradient through that judgment (\S\ref{sec:method}). In place of GRPO's scalar advantage, SDAE computes a group-relative advantage along each of $K$ subtask axes (addressing the across-competency facet, and recovering scalar GRPO when $K{=}1$) and shapes per-token credit around the reflector-identified pivot for each competency (addressing the across-position facet). The per-token gradient now varies both across competencies and across positions in the same rollout. In our motivating example, SDAE would assign a positive per-subtask advantage to the second rollout's object-gathering tokens (which succeeded relative to its group) while assigning a negative advantage to its scientific-reasoning tokens (which failed); scalar GRPO cannot make this distinction.

We evaluate RLDS on four agentic benchmarks (FrozenLake, HotpotQA, ScienceWorld, and DeepResearch) and treat the experiments as a test of a prediction our training-time diagnostics (\S\ref{sec:analysis}) make: gains should scale with how much the per-subtask signal varies within a group, and with how widely the per-competency pivots spread across the trajectory. These are the two axes of variation that scalar GRPO with uniform per-token broadcast averages away.

\section{Related Work}
\label{sec:bg}

RLDS sits at the intersection of reflection-based RL and token/step-level credit assignment. We briefly introduce both lines of work.

\subsection{Reflection-Based RL}
\label{sec:bg:reflection}

Self-reflection-based fine-tuning strategies allow the language model to reason using natural language in order to better correct potential mistakes \citep{shinn2023, bensal2025reflect, renze2024self}. Our method draws on two recent reflection-based approaches, ERL \citep{shi2026} and R$^3$L \citep{shi2026r3l}.

ERL \citep{shi2026} adds a reflect-then-retry phase inside each PPO \citep{schulman2017proximal} step. After an initial rollout, the agent emits a replacement system prompt inside \texttt{<improved\_prompt>} tags conditioned on the failed trajectory and a cross-episode memory of past rewrite attempts; a second rollout is drawn under that rewritten prompt; both attempts are optimized through the same group-relative objective; and a parallel SFT distillation loss trains the policy to reproduce the successful retry response given the \emph{original} (unrewritten) prompt, internalizing retry-quality behavior into the base policy so the deployed model does not depend on reflection at inference time. We adopt this retry-distillation channel directly as part of our training recipe and do not claim novelty for it.

R$^3$L \citep{shi2026r3l} keeps the original system prompt fixed and replaces the rewrite with a structured JSON reflection report containing a root-cause analysis, an improvement suggestion, and a \texttt{retry\_from\_step} index. The retry resumes from that pivot conditioned on the reflection, and the gradient is shaped through three additional levers: a pivotal credit mask that zeros gradients on the shared prefix between the original attempt and the retry, an OPMD-style baseline that replaces the GRPO \citep{shao2024deepseekmath} group baseline with a retry-conditional one, and positive amplification that upweights the advantages of successful retries to prevent failure-dominated entropy collapse. R$^3$L also introduces an auxiliary SFT term that does NLL on the reflection text itself given its conditioning context, complementing ERL's response-side distillation; we adopt this term and discuss it alongside ERL's retry distillation in \S\ref{sec:method:setup}. We keep the report, the pivot mask, and the reflection-NLL term, and drop the OPMD baseline and positive amplification.

\subsection{Token- and Step-Level Credit Assignment}
\label{sec:bg:credit}

A separate line of work redistributes credit inside the policy gradient without reflection. GTPO and GRPO-S \citep{tan2025gtpo} reshape the GRPO advantage by per-token policy entropy, concentrating credit on tokens likely to carry decision-relevant information; GRPO-$\lambda$ \citep{parthasarathi2025grpolambda} uses eligibility traces and TD-style propagation to flow terminal reward backward through the trajectory. These methods \emph{distribute} the scalar reward but treat entropy as a proxy for where credit should land, not a diagnosis of what should be reinforced.

A complementary direction trains a verifier for step-level supervision. Process Reward Models \citep{lightman2024} score each step of a reasoning trace; AgentPRM \citep{choudhury2025agentprm} extends this to agentic rollouts via Monte Carlo lookahead and a lightweight actor--critic update. PRMs factor reward cleanly but require human annotation or a trained verifier, and they factorize along the temporal step axis rather than competency dimensions: distinct failures on the same step, e.g., a wrong tool argument and a wrong reasoning premise, collapse into a single score.

\section{Reinforcement Learning with Decomposed Subtasks}
\label{sec:method}

\begin{figure}[h]
\centering
\vspace{-0.2cm}
\includegraphics[width=\textwidth]{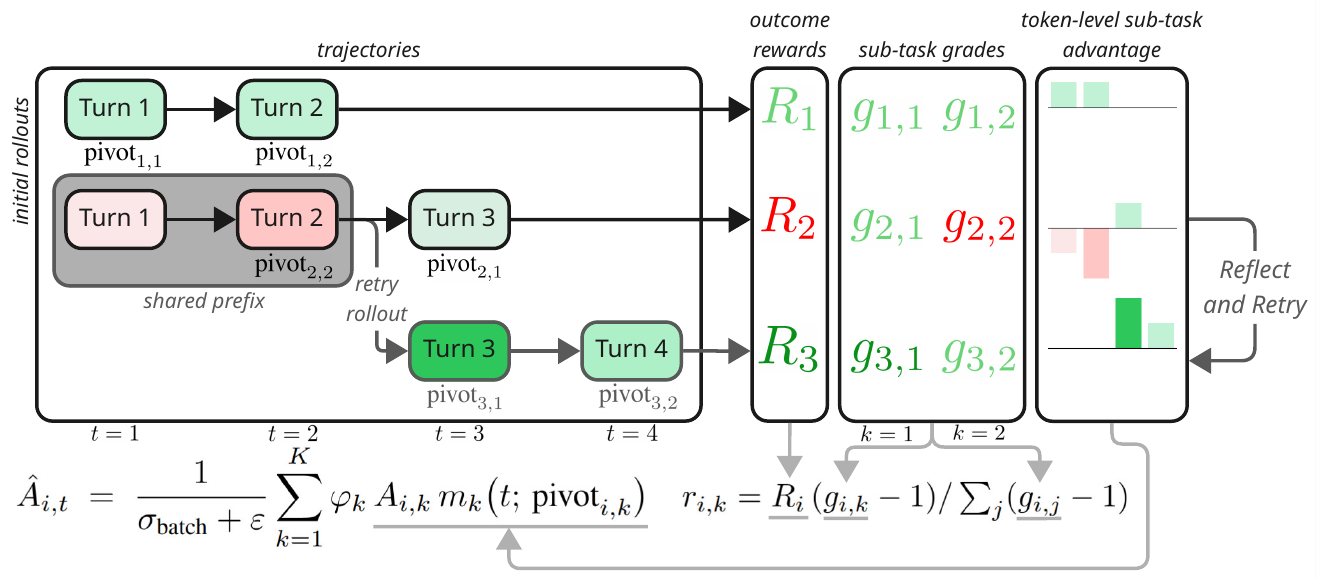}
\vspace{-0.5cm}
\caption{Visualization of SDAE for a single rollout with $G=2$, $K=2$ and $\mathcal{R}=1$. We compute the final token-wise advantage $\hat A_{i,t}$ using reward values $R$ from the environment, distributed among subtask grades $g_{i,k}$ from the policy model and credited to tokens using a gaussian kernel around pivot points $\text{pivot}_{i,k}$. Red and green hues indicate the relative values of rewards, subtask grades and token-wise advantage.}
\label{fig:overview}
\end{figure}

We begin by motivating our method in \S\ref{sec:motivation}. We visualize the pipeline end-to-end in Figure~\ref{fig:overview} and Algorithm~\ref{alg:rlds}. We adopt the reflect-retry rollout from R$^3$L~\citep{shi2026r3l}; \S\ref{sec:method:setup} describes the operational details we rely on, \S\ref{sec:method:sdae} defines the three-stage SDAE advantage at the core of RLDS, and \S\ref{sec:method:objective} states the combined objective. 

\subsection{Motivation}
\label{sec:motivation}

Reflection-based methods generate trajectories that diagnose \emph{what} failed in natural language but feed a scalar reward back to the optimizer. Token- and step-level credit-assignment methods redistribute reward inside a trajectory but treat the trajectory as opaque text, they decide \emph{where} credit should land through entropy heuristics or learned step-level verifiers, not through an explicit decomposition of competencies. The reflection identifies which subtask each part of the trajectory was working on, the pivot localizes the decisive turn, and the retry exposes a counterfactual under which per subtask improvements can be measured. The contribution of this paper is RLDS, the recipe that consumes that signal: SDAE, the advantage estimator at its core, together with the mechanism we define to distribute those rewards across the trajectory.

\subsection{Preliminaries: Subtask Self-Grading on Reflect-Retry Rollouts}
\label{sec:method:setup}

Each task is annotated with a fixed decomposition of $K$ named subtasks $\{s_k\}_{k=1}^{K}$ with importance weights $\{\varphi_k\}$ summing to one, held constant throughout training and inference (per-task taxonomies in Appendix~\ref{app:taxonomies}).

\paragraph{Reflect-retry rollout.} Each training step, $\pi_\theta$ draws $G$ initial rollouts $\{y^{(1)}_i\}_{i=1}^{G}$ from a task and observes a scalar trajectory reward $R_i$ for each. The same policy then self-grades each rollout under the taxonomy, emitting a per-subtask Likert grade $g_{i,k} \in \{1, \ldots, 5\}$ and a per-subtask pivot step $\text{pivot}_{i,k}$, the turn the grader identifies as the point at which subtask $k$'s outcome was determined. The bottom-$\mathcal{R}$ failures are selected as reflection candidates; for each, $\pi_\theta$ emits a structured JSON reflection report and retries from \texttt{retry\_from\_step} to produce a corrected rollout $y^{(2)}_j$, which is itself self-graded the same way. Originals and retries are folded into a single GRPO group of size $G{+}\mathcal{R}$ and treated uniformly thereafter. The rollout follows R$^3$L~\citep{shi2026r3l}; the prompts $\pi_\theta$ are conditioned on as grader and reflector are reproduced verbatim in Appendix~\ref{app:prompts}.

\paragraph{$\pi_\theta$ is actor, grader, and reflector.} The same policy plays all three generative roles here, namely actor (rollouts and retries), grader, and reflector, rather than offloading grading or reflection to an external judge. The auxiliary supervision pairs introduced below supervise $\pi_\theta$ in exactly the grader and reflector roles, so improvements on those skills compound with policy training; keeping every generative role on-policy also minimizes the distribution shift between sampled and trained tokens. The flip side is that the actor and grader share parameters, so we discipline the self-grader by requiring that subtask grades only \emph{redistribute} the trajectory reward $R_i$ across the $K$ axes; they never scale it. By construction, $\sum_k r_{i,k} = R_i$ for every rollout: the environment owns reward magnitude and the grader owns its allocation across competencies, removing the moral hazard a self-grading judge would otherwise introduce.

\paragraph{Auxiliary supervision pairs.} Beyond the GRPO group, the same rollout produces two off-policy distillation pairs per successful retry: a (reflection, conditioning context) pair for R$^3$L's reflection-NLL term~\citep{shi2026r3l} and an (original prompt, retry response) pair for ERL's retry-distillation term~\citep{shi2026}; both channels and their motivations are introduced in \S\ref{sec:bg:reflection}. We retain the top-$K$ pairs per step ranked by retry reward delta, weight each pair by its delta so high-impact reflections dominate, and sum the two NLL terms into the auxiliary loss $\mathcal{L}_{\text{aux}}$ that enters the full objective in \S\ref{sec:method:objective}.

\subsection{Subtask-Decomposed Advantage Estimation}
\label{sec:method:sdae}

\paragraph{Stage 1. Group-relative per-subtask advantages.} We split each trajectory's reward into per-subtask shares using a zero-anchored proportion of the grades, $r_{i,k} = R_i\,(g_{i,k} - 1) / \sum_{j} (g_{i,j} - 1)$ (uniform $r_{i,k} = R_i/K$ when every $g_{i,j} = 1$); the $-1$ matches the rubric's grade-1-as-failure semantics so a critically-failed dimension contributes no share before baselining. Within each GRPO group of size $G{+}\mathcal{R}$ we then compute a subtask-specific mean and the per-subtask advantage:
\begin{equation}
  \bar r_k \;=\; \frac{1}{G+\mathcal{R}}\sum_{i=1}^{G+\mathcal{R}} r_{i,k}, \qquad A_{i,k} \;=\; r_{i,k} - \bar r_k.
  \label{eq:stage1}
\end{equation}
whereas in vanilla GRPO, the scalar baseline subtracts the group mean trajectory reward from each row, so two rollouts with different competency profiles but the same total reward look identical to the optimizer. Baselining each subtask axis separately exposes the dimensions on which a rollout was above or below the group along that axis, even when its trajectory reward is at the mean. The sign of $A_{i,k}$ now records, per subtask, whether the rollout was above or below average on that competency.

\paragraph{Stage 2. Importance-weighted sum and batch normalization.} The per-subtask advantages are recombined into a row-level scalar under the static importance weights and normalized at the batch level:
\begin{equation}
  A_i \;=\; \sum_{k=1}^{K} \varphi_k\, A_{i,k}, \qquad \hat A_i \;=\; \frac{A_i}{\sigma_{\text{batch}} + \varepsilon},
  \label{eq:stage2}
\end{equation}
where $\sigma_{\text{batch}}$ is the standard deviation of $\{A_i\}$ across all decomposed rows in the batch. Batch-level normalization (rather than group-level) absorbs scale differences across tasks when training mixes them and keeps the optimizer well-conditioned. Groups for which no per-subtask grades were emitted (e.g., a grader-parse failure) fall back to vanilla GRPO advantages on the trajectory reward, so SDAE degrades gracefully on missing data. We log three diagnostics per step that characterize how much of the multi-dimensional signal SDAE actually recovers: sign-disagreement rate (the fraction of rollouts whose per-subtask advantages disagree in sign within the group), GRPO correlation (Pearson correlation between the $\varphi$-weighted SDAE advantage and the scalar GRPO advantage on the same rollouts), and token-level advantage std (the per-row spread of per-token advantages once Stage~3's pivot kernel is applied). All three are analyzed in \S\ref{sec:analysis:rlds-metrics}.

\paragraph{Stage 3. Token application.} Stage 2 returns a row-level scalar; before it enters the PPO surrogate, we expand it back into a per-token advantage. The default GRPO move is to broadcast $\hat A_i$ uniformly over the response tokens of rollout $i$. SDAE supports that broadcast as a fallback, but when pivot steps are available we instead apply each subtask's advantage through its own per-token kernel centered on the turn the grader marked as decisive:
\begin{equation}
  \hat A_{i,t} \;=\; \frac{1}{\sigma_{\text{batch}} + \varepsilon}\sum_{k=1}^{K} \varphi_k\, A_{i,k}\, m_k\bigl(t;\, \text{pivot}_{i,k}\bigr),
  \label{eq:stage3}
\end{equation}
where $\sigma_{\text{batch}}$ is the same denominator as in Stage 2 and $m_k(t;\, \text{pivot}_{i,k})$ is a smooth kernel in turn-units centered on $\text{pivot}_{i,k}$, normalized to mean one over the response tokens of rollout $i$ so that the kernel \emph{redistributes} credit across positions without changing the per-row total. The shared $\sigma_{\text{batch}}$ keeps Stage 2's broadcast and Stage 3's per-token expansion on the same scale: averaging $\hat A_{i,t}$ over response tokens recovers $\hat A_i$. We use a Gaussian kernel in all reported experiments. Pivot masks compose multiplicatively with R$^3$L's global retry-step mask, which zeros gradients on the prefix shared between an original attempt and its retry, so on retry rows the gradient concentrates on the diverging suffix conditioned on the reflection.

\subsection{Combined Objective}
\label{sec:method:objective}

The full training loss combines a PPO surrogate evaluated against the SDAE per-token advantage with the auxiliary distillation loss $\mathcal{L}_{\text{aux}}$ defined in \S\ref{sec:method:setup}:
\begin{equation}
  \mathcal{L}_{\text{total}} \;=\; \mathcal{L}_{\text{PPO}}\bigl(\hat A^{\text{SDAE}}_{i,t}\bigr) \;+\; \mathcal{L}_{\text{aux}},
  \label{eq:combined}
\end{equation}
where $\mathcal{L}_{\text{PPO}}$ is the standard clipped surrogate with a per-token KL penalty against a frozen reference and $\mathcal{L}_{\text{aux}}$ is the auxiliary distillation loss introduced in \S\ref{sec:method:setup}, weighted per-pair internally by retry reward delta as described there.

\begin{algorithm}[H]
  \caption{RLDS training step.}
  \label{alg:rlds}
  \begin{algorithmic}[1]
    \Require task $x$, subtask taxonomy $\{s_k\}_{k=1}^{K}$ with importance weights $\{\varphi_k\}$, group size $G$, retry budget $\mathcal{R}$, policy $\pi_\theta$
    \State Sample initial rollouts $\{y^{(1)}_i\}_{i=1}^{G} \sim \pi_\theta(\cdot \mid x)$ and observe trajectory rewards $\{R_i\}$
    \State Self-grade each $y^{(1)}_i$ via $\pi_\theta$: per-subtask Likert grade $g_{i,k} \in \{1,\ldots,5\}$ and pivot step $\text{pivot}_{i,k}$
    \State Select bottom-$\mathcal{R}$ failures by reward as reflection candidates
    \State For each candidate, emit a structured reflection report and retry from \texttt{retry\_from\_step} to produce $\{y^{(2)}_j\}_{j=1}^{\mathcal{R}}$
    \State Self-grade retries; fold originals and retries into one GRPO group of size $G{+}\mathcal{R}$
    \Statex \hrulefill
    \Statex \textbf{SDAE advantage}:
    \State Decompose: $r_{i,k} \gets p_{i,k} R_i$, where $p_{i,k} = (g_{i,k} - 1) / \sum_j (g_{i,j} - 1)$ \ (uniform $1/K$ if all $g_{i,j} = 1$)
    \State \textbf{Stage 1.} \ Group baseline: $A_{i,k} \gets r_{i,k} - \bar r_k$, where $\bar r_k = \tfrac{1}{G+\mathcal{R}} \sum_{i=1}^{G+\mathcal{R}} r_{i,k}$
    \State \textbf{Stage 2.} \ Importance-weighted sum + batch norm: $A_i \gets \sum_k \varphi_k A_{i,k}$, \ $\hat A_i \gets A_i / (\sigma_{\text{batch}} + \varepsilon)$
    \State \textbf{Stage 3.} \ Token application: $\hat A_{i,t} \gets \tfrac{1}{\sigma_{\text{batch}} + \varepsilon}\sum_k \varphi_k A_{i,k} \cdot m_{i,k,t}$, where $m_{i,k,t}$ is a kernel concentrated around $\text{pivot}_{i,k}$
    \Statex \hrulefill
    \State PPO/GRPO policy update with $\hat A_{i,t}$ and the auxiliary distillation loss $\mathcal{L}_{\text{aux}}$ (single backward pass)
  \end{algorithmic}
\end{algorithm}

\section{Experiments}
  \label{sec:exp}

We evaluate RLDS against scalar GRPO and the reflect-retry baseline (Algorithm~\ref{alg:rlds}, lines 1--5, with vanilla GRPO advantages) on four agentic benchmarks chosen to span the subtask heterogeneity axis. RLVR uses group size $G{=}10$; reflection-enabled conditions use $G{=}6$ originals plus up to $\mathcal{R}{=}3$ retries, folded into a combined GRPO group of size $G{+}\mathcal{R}$ to match the per-step token budget. Subtask taxonomies, importance weights $\varphi_k$, reward formulas, and tool specifications are deferred to Appendices~\ref{app:taxonomies}--\ref{app:benchmarks}.

\textbf{FrozenLake.} Grid-world navigation, no tools, 10 turns, sparse outcome reward. \textbf{HotpotQA.} Multi-hop QA with a single \texttt{search} tool over Wikipedia, 5 turns, F1-based reward; retrieval follows Search-R1~\citep{jin2025searchr1}. \textbf{ScienceWorld}~\citep{wang2022scienceworld}. Text-based interactive science, 30 procedural tasks, no tools, 30 turns, simulator partial credit. \textbf{DeepResearch.} Long-form grounded research with four tools, 10 turns, LLM-judged rubric plus deterministic formatting scores; setup follows DR Tulu~\citep{shao2025drtulu}.

\section{Analysis}
\label{sec:analysis}

\begin{figure}[h]
\centering
\includegraphics[width=\textwidth]{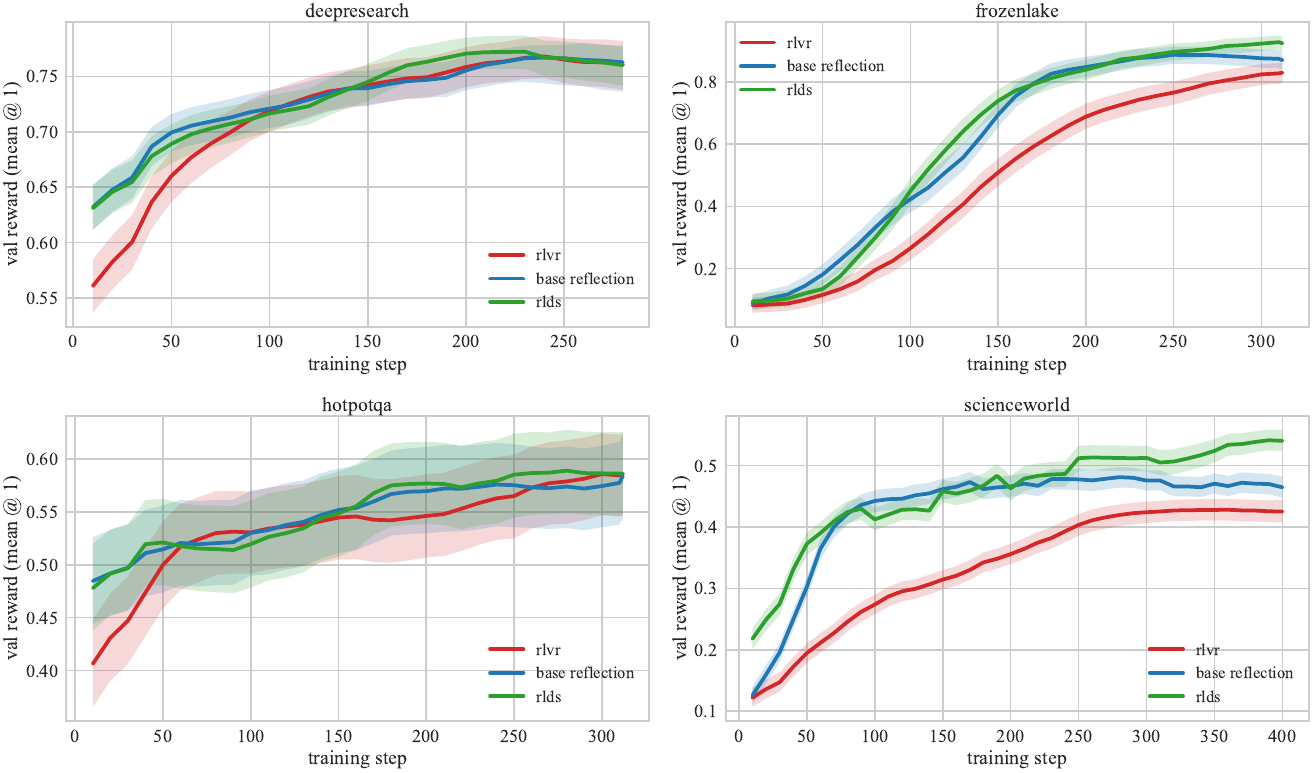}
\vspace{-0.3cm}
\caption{Best validation reward across tasks for scalar GRPO (\textsc{rlvr}), the reflect-retry baseline (\textsc{base reflection}), and \textsc{rlds}. Shaded bands are 95\% non-parametric bootstrap confidence intervals on the per-checkpoint validation mean, obtained by resampling the validation set with replacement ($B{=}1000$); they reflect evaluation-set sampling noise on a fixed checkpoint and do not bound training-trajectory variance across seeds.}
\label{fig:val-reward-best}
\end{figure}

RLDS's gain over GRPO scales with how \emph{process-heavy} and \emph{long-horizon} a task is: the more subtask structure spreads across many turns, the more credit decomposition recovers along the within-group variance and pivot-spread axes \S\ref{sec:intro} predicted gains would track. ScienceWorld coordinates navigation, manipulation, and procedural reasoning over 30-turn rollouts; FrozenLake interleaves hazard avoidance and dynamics inference turn by turn. The per-subtask decomposition recovers heterogeneous signal that GRPO averages flat. HotpotQA and DeepResearch carry minimal process content over short horizons; RLDS figures this out quickly, per-subtask variance reduces, pivots and retries land low-impact, and the SDAE advantage regresses to scalar RLVR. Figure~\ref{fig:val-reward-best} reports best validation reward by task with the largest gain on ScienceWorld (+11.5 points over RLVR).

\subsection{SDAE Metrics}
\label{sec:analysis:rlds-metrics}

SDAE turns the scalar advantage into a per-token tensor through a three-stage pipeline: the rollout has to carry heterogeneous per-subtask signal at all, the $\varphi$-weighted decomposition has to then change which rollouts get the strongest gradient, and pivot masking finally has to redistribute that credit across tokens within a rollout. We emit one diagnostic per stage; a task that fails any of them collapses back to scalar broadcast at that point and SDAE has nothing to add downstream.

\begin{wrapfigure}[13]{r}{0.42\textwidth}
\centering
\includegraphics[width=0.4\textwidth]{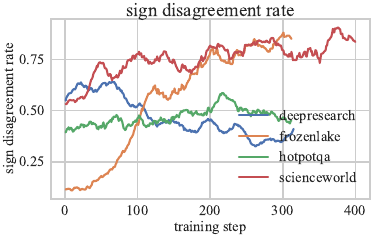}
\caption{Sign-disagreement rate.}
\label{fig:rlds-sign-disagreement}
\end{wrapfigure}
\paragraph{Stage 1: Is there heterogeneous signal? Sign-disagreement rate.} Fraction of GRPO groups whose per-subtask advantages mix signs within a trajectory. A disagreeing row is precisely where scalar GRPO averages flat: the policy did well on competency A and badly on B, the trajectory reward folds them into one number, while RLDS preserves the signs through SDAE so each token gets gradient pushing toward the competency it served. Tasks with persistently high disagreement (Figure~\ref{fig:rlds-sign-disagreement}, ScienceWorld) clear this first gate and leave the most for the later stages to act on; tasks where subtasks move together fail it and SDAE has nothing to recover beyond GRPO's verdict regardless of how the rest of the pipeline is configured. Disagreement also serves as an early warning during training: if it collapses mid-run while reward keeps improving, the policy has converged onto a single-competency strategy and the decomposition is silently degenerating to a scalar.

\begin{wrapfigure}[13]{r}{0.42\textwidth}
\centering
\includegraphics[width=0.4\textwidth]{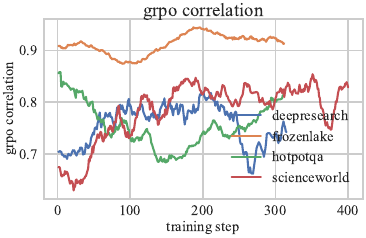}
\caption{GRPO correlation.}
\label{fig:rlds-grpo-corr}
\end{wrapfigure}
\paragraph{Stage 2: Does decomposition reweight rollouts? GRPO correlation.} Pearson correlation between the per-row $\varphi$-weighted SDAE advantage (before pivot masking) and the scalar GRPO advantage on the same rollouts. Given that stage 1 has confirmed heterogeneous signal exists, this asks whether the decomposition step survives $\varphi$-weighting or cancels back to the scalar. A correlation near 1.0 means the per-subtask signal cancels and the decomposition is informationally redundant at the rollout level; a correlation closer to 0.6 (Figure~\ref{fig:rlds-grpo-corr}) means RLDS is reweighting which rollouts get the strongest gradient on a non-trivial fraction of the batch.

\begin{wrapfigure}[13]{r}{0.42\textwidth}
\centering
\includegraphics[width=0.4\textwidth]{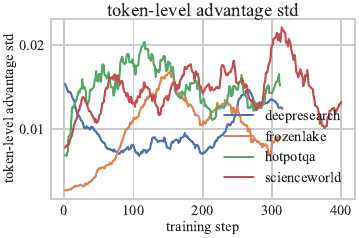}
\caption{Token-level advantage std.}
\label{fig:rlds-token-std}
\end{wrapfigure}
\paragraph{Stage 3: Does pivot masking redistribute credit within a rollout? Token-level advantage std.} For each row with pivot data, the standard deviation of per-token advantage values across response-mask-active positions, averaged across the batch. Conditional on stages 1 and 2 having fired, this isolates the spatial-shaping path: scalar broadcast produces zero by construction, and a non-zero value arises only when the per-subtask pivot masks land on different turn windows \emph{and} the per-subtask advantages differ in sign or magnitude. Figure~\ref{fig:rlds-token-std} confirms the path is active on the heterogeneous tasks; on HotpotQA, tokens near the late-trajectory \texttt{answer\_grounding} pivot get visibly different credit than tokens near the early-trajectory \texttt{search\_planning} pivot.

The cascade lines up with the reward lifts \S\ref{sec:analysis:fit} reports: ScienceWorld clears all three gates and posts the largest gain, while HotpotQA and DeepResearch wash out at stage 1 over their short, process-light horizons and the SDAE advantage regresses toward scalar RLVR --- which is the behavior we want when the rollout is not producing competency-tagged signal worth decomposing.

\subsection{Compute Profile}
\label{sec:analysis:fit}

Each RLDS step samples $G$ originals, prompts the policy to grade and reflect on the bottom-reward ones, samples up to $\mathcal{R}$ retries, and grades each retry per-subtask --- a fixed per-step overhead on top of the rollout. The interesting question is how this overhead scales with horizon, because the same machinery that costs the most on short tasks pays for itself on long ones.

\paragraph{Compute inverts at long horizons.} Short-horizon step time roughly doubles ($2.4{\times}$/$2.3{\times}$/$1.6{\times}$ on FrozenLake/HotpotQA/DeepResearch), driven by reflection and grading. At long horizons retries resample only the suffix from the pivot (retry-to-original gen ratio $0.76$/$0.48$/$0.22$ on FrozenLake/HotpotQA/ScienceWorld) and the reflection-plus-grading overhead shrinks once each rollout spans $\sim$$4$\,K tokens over $\sim$$33$ turns; ScienceWorld lands at $0.89{\times}$ step-time despite strictly more work per trajectory. Reward gains are reported at fixed step count; a budget-matched comparison is left to future work.

\begin{table}[h]
\centering
\small
\caption{Per-task compute and reward lift. \emph{Avg turns}: mean agent--environment turns per episode. \emph{Gen}: rollout time ($G$ originals for RLVR; plus $\mathcal{R}$ retries for RLDS). \emph{Judge}: per-subtask grading time (RLDS only). \emph{Step}: training-step wall-clock. \emph{$\Delta$ Step} $= (t_{\text{RLDS}} - t_{\text{RLVR}}) / t_{\text{RLVR}}$, negative if RLDS is faster.}
\label{tab:compute-profile}
\setlength{\tabcolsep}{4pt}
\begin{tabular}{lrrrrrrrr}
\toprule
& & \multicolumn{2}{c}{Gen (s)} & Judge (s) & \multicolumn{2}{c}{Step (s)} & & \\
\cmidrule(lr){3-4} \cmidrule(lr){5-5} \cmidrule(lr){6-7}
Task & Avg turns & RLVR & RLDS & RLDS & RLVR & RLDS & $\Delta$ Step & Lift \\
\midrule
HotpotQA      &  5 &  28.1 &  64.5 & 25.8 &  69.5 & 157.0 & $+125.9\%$ & $+1.9\% \pm 5.1\%$  \\
DeepResearch  &  8 & 118.0 & 162.9 & 24.3 & 163.4 & 267.9 & $+64.0\%$  & $+0.7\% \pm 1.7\%$    \\
FrozenLake    & 10 &  27.9 &  41.8 & 22.2 &  49.8 & 119.0 & $+139.0\%$ & $\bm{+11.7\% \pm 3.8\%}$ \\
ScienceWorld  & 38 & 221.5 & 170.3 & 25.3 & 294.4 & 262.2 & $\bm{-10.9\%}$  & $\bm{+26.7\% \pm 4.8\%}$ \\
\bottomrule
\end{tabular}
\end{table}

\subsection{Subtask Metrics}
\label{sec:analysis:subtask-metrics}

The step-level diagnostics in \S\ref{sec:analysis:rlds-metrics} compress every subtask into one number per step. They tell us \emph{whether} RLDS recovers multi-competency signal but not \emph{which} subtask is responsible. We expose two per-subtask readouts: \textbf{retry grade deltas} (mean change in grader score from the failed original to its reflect-retry) and \textbf{per-subtask pivot step} (average turn where the pivot mask lands). They answer the questions: did reflection help \emph{this} competency and where does it sit in the trajectory. DeepResearch and HotpotQA analyses are available in the appendix.

\paragraph{FrozenLake.} Figure~\ref{fig:subtasks-frozenlake} shows retry grade deltas are led by \texttt{hazard\_identification} (near $+1.8$) and \texttt{path\_planning}. \texttt{dead\_end\_detection} picks up sharply at the end of training as the remaining failure cases are tricky maps where the agent could get stuck. Pivot steps means intuitively place the \texttt{dead\_end\_detection} deeper in the trajectory, while \texttt{hazard\_identification} and \texttt{path\_planning} are activities that happen towards the beginning.

\begin{figure}[h]
\centering
\includegraphics[width=\textwidth]{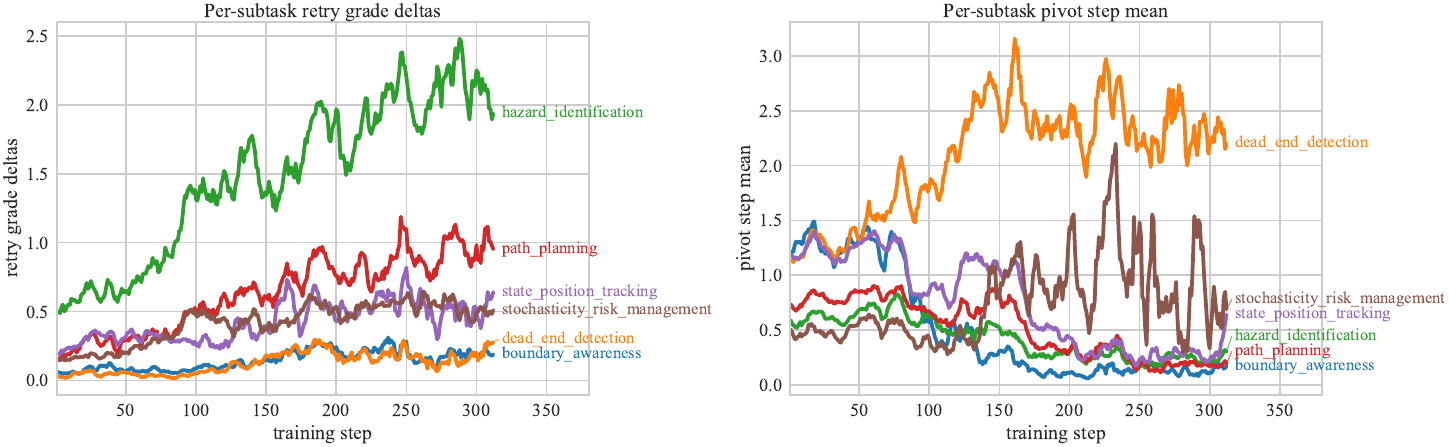}
\caption{FrozenLake per-subtask metrics over training: retry grade deltas and pivot step.}
\label{fig:subtasks-frozenlake}
\end{figure}

\begin{figure}[h]
\centering
\includegraphics[width=\textwidth]{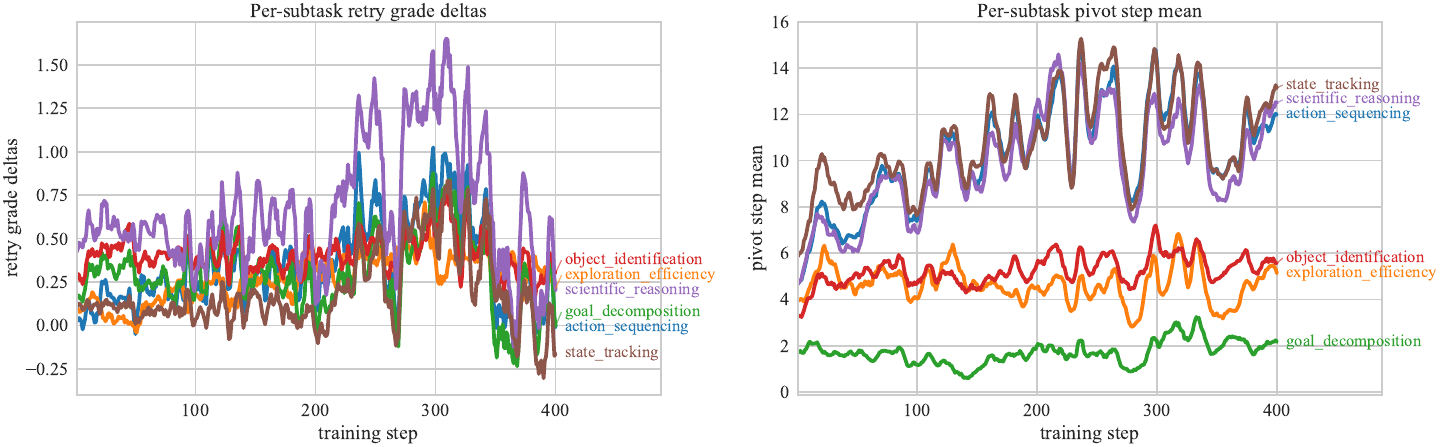}
\caption{ScienceWorld per-subtask metrics over training: retry grade deltas and pivot step.}
\label{fig:subtasks-scienceworld}
\end{figure}

\paragraph{ScienceWorld.} ScienceWorld is the long-horizon stress test (30 turns, six subtasks) that demonstrates how subtask importance can shift over time. Figure~\ref{fig:subtasks-scienceworld} shows that around step 200 the model has figured out how to interact with the environment and what becomes more important is the \texttt{scientific\_reasoning} competency. Once objects are collected this is the subtask that allows the model to compose experiments to answer the scientific questions at hand. Pivot steps fan across the 30-turn budget: \texttt{goal\_decomposition} at $\approx 2$, \texttt{exploration\_efficiency} and \texttt{object\_identification} mid ($\approx 5$), and \texttt{action\_sequencing}, \texttt{scientific\_reasoning}, \texttt{state\_tracking} late ($\approx 11$--$13$). The temporal range is the widest of any task.

\section{Limitations}
\label{sec:limitations}

\paragraph{Defining subtasks.}
The taxonomy is fixed per task and took several iterations per environment. We recommend aligning subtasks with \emph{temporal stages} of the trajectory (e.g., HotpotQA's \texttt{search\_planning} $\to$ \texttt{search\_iteration} $\to$ \texttt{evidence\_extraction} $\to$ \texttt{answer\_grounding}) rather than orthogonal competencies (``reasoning'' vs.\ ``tool use'' vs.\ ``self-correction''); the latter produces graders whose pivot steps collapse onto the same turn, leaving the SDAE advantage doing scalar broadcast in disguise.

\paragraph{Static subtask weights.}
The importance weights $\varphi_k$ are static per task, and tuning them falls on the practitioner. Since we already log per-subtask retry grade deltas, a natural extension is to make $\varphi_k$ dynamic: place a Dirichlet prior over the weights and update its concentration from the observed deltas, so subtasks that consistently benefit from reflection are upweighted online while saturated ones decay.

\section{Conclusion}
\label{sec:conclusion}

Reflect-retry rollouts already produce a rich diagnosis: a competency-tagged judgment of what failed and which turn determined each failure. Scalar GRPO collapses that diagnosis into a single number per trajectory and forces the optimizer to average credit over competencies it cannot tell apart and across positions where they never fired. RLDS consumes the diagnosis the rollout was already producing through the SDAE advantage: group-relative per-subtask baselines preserve the directional disagreements scalar GRPO would average flat, and per-subtask pivot masks reshape credit across response tokens around the turn the grader marked decisive. The prediction that fell out of this construction, that gains should track how much the per-subtask signal varies within a group and how widely the per-competency pivots spread across the trajectory, is what the four benchmarks were chosen to probe, and it is what the analysis bears out. ScienceWorld carries within-group sign disagreement throughout training and fans its per-competency pivots from turn $\approx 2$ to $\approx 13$ across a 30-turn budget; it posts a $+26.7\%$ lift and, because retries resample only the suffix from the pivot, costs $0.89{\times}$ a scalar GRPO step despite the extra reflection and grading. FrozenLake carries the same heterogeneity at a shorter horizon and gains $+11.7\%$. HotpotQA and DeepResearch fail the within-group variance gate over their short, process-light horizons; SDAE regresses toward scalar RLVR by construction, and the lifts sit in noise. The cascade of SDAE diagnostics in \S\ref{sec:analysis:rlds-metrics}, sign-disagreement rate, GRPO correlation, and token-level advantage std, locates exactly where each task lives on this gradient, and the per-subtask retry deltas and pivot steps (\S\ref{sec:analysis:subtask-metrics}) give the practitioner a continuous read on training. Subtask decomposition is the right primitive for multi-competency RL because the rollout produces more information than the scalar can carry. When the rollout already produces competency-tagged signal, the advantage estimator should consume it.

\bibliography{references}

\appendix

\section{5.3 Subtask Metrics continued}

\begin{figure}[h]
\centering
\includegraphics[width=\textwidth]{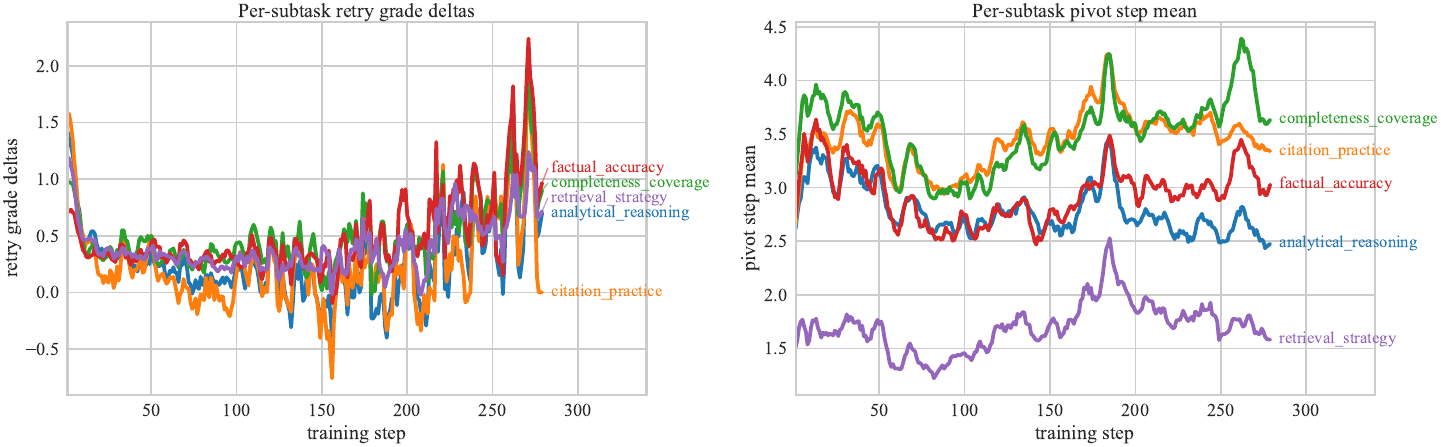}
\caption{DeepResearch per-subtask metrics over training: retry grade deltas and pivot step.}
\label{fig:subtasks-deepresearch}
\end{figure}

\paragraph{DeepResearch.} DeepResearch's five-skill taxonomy (\texttt{retrieval\_strategy}, \texttt{analytical\_reasoning}, \texttt{factual\_accuracy}, \texttt{completeness\_coverage}, \texttt{citation\_practice}) maps onto the natural temporal stages of a research trajectory, and Figure~\ref{fig:subtasks-deepresearch} shows the pivot steps separating accordingly within the 10-turn budget: \texttt{retrieval\_strategy} sits earliest at $\approx 1.5$, \texttt{analytical\_reasoning} and \texttt{factual\_accuracy} in the middle at $\approx 2.5$--$3$, and \texttt{completeness\_coverage} and \texttt{citation\_practice} latest at $\approx 3.5$--$4$ --- search first, synthesize and state facts in the middle, cover and cite at the end. Retry grade deltas compress after the first ${\sim}50$ steps and then climb sharply past step $200$, led by \texttt{factual\_accuracy} (peaks $\approx 2.0$); \texttt{citation\_practice} hovers near zero throughout, consistent with the deterministic formatting score already disciplining that channel and leaving little for reflection to add.

\begin{figure}[h]
\centering
\includegraphics[width=\textwidth]{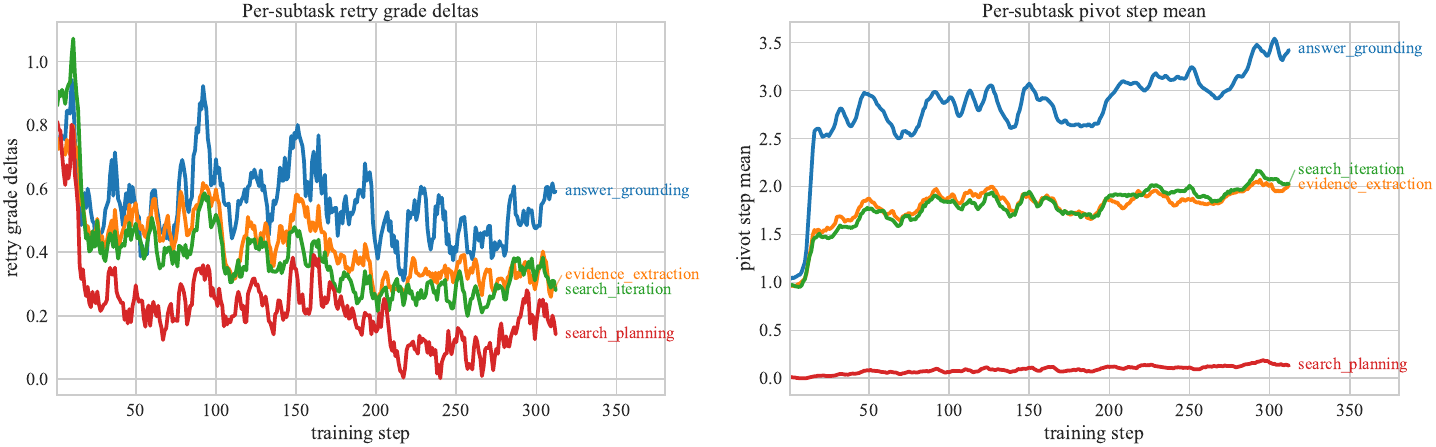}
\caption{HotpotQA per-subtask metrics over training: retry grade deltas and pivot step.}
\label{fig:subtasks-hotpotqa}
\end{figure}

\paragraph{HotpotQA.} Figure~\ref{fig:subtasks-hotpotqa} visualizes HotpotQA's four-skill taxonomy (\texttt{search\_planning}, \texttt{search\_iteration}, \texttt{evidence\_extraction}, \texttt{answer\_grounding}) gives the cleanest per subtask picture in the suite from a temporal perspective. The retry grade deltas fade quickly and settle into a small positive gain. This also maps to initial over-performance on the validation reward as shown in Figure~\ref{fig:val-reward-best} that fades by step 50.

\section{Training Runs}
\label{app:training-runs}

\paragraph{Shared training recipe.} All runs fine-tune Qwen3-4B-Instruct-2507 with GRPO via verl~+ SGLang. The optimizer is AdamW with a constant learning rate of $1{\times}10^{-6}$, fixed KL coefficient $0.001$ against a frozen reference, PPO clip ratio $0.28$, one PPO epoch per step, and entropy coefficient $0$. Each run does one epoch over the task's train split with validation every 10 update steps; we report best validation reward across the run. The actor uses FSDP2 with the SGLang rollout engine and the verl multi-turn agent loop in Hermes tool-call format. Each run uses one node with $8{\times}$H100 80\,GB.

\paragraph{Per-task hyperparameters and run statistics.} Table~\ref{tab:hparams} reports, for each task, the dataset sizes, the runtime knobs that differ from the shared recipe (max prompt/response, max turns, temperature, group size, batch size), the total number of update steps, the median wall-clock per step and per run, the hardware, and the volume of external-API traffic that the rollout loop generates. Step times and step counts are pulled directly from MLflow (\texttt{perf/time\_per\_step}); all other rows are pinned configuration. Tool-call and judge-call volumes are reported per rollout (mean) and totalled per RLDS run; only DeepResearch has paid external dependencies.

\begin{table}[h]
\centering
\footnotesize
\caption{Per-task training run statistics. Slash-separated cells report \emph{RLVR / RLDS} where the two conditions diverge. \emph{Tool calls / rollout} are MLflow-logged per-rollout means at training-time temperature; \emph{Total tool calls / run} aggregate originals + retries across the full RLDS run. Cache hit rates on Serper-backed tools sit at $1$--$5\%$ during training (the cache is most useful across seeds, not within a single run). Dollar costs are deliberately omitted because they depend on Serper / OpenAI contract pricing tier; raw call volumes are reported instead so cost can be computed against the reader's own pricing.}
\label{tab:hparams}
\setlength{\tabcolsep}{4pt}
\begin{tabular}{lrrrr}
\toprule
 & FrozenLake & HotpotQA & ScienceWorld & DeepResearch \\
\midrule
\multicolumn{5}{l}{\emph{Data}} \\
Train examples                        & 20{,}000 & 20{,}000 & 1{,}796   & 6{,}480 \\
Validation examples                   & 500      & 500      & 887       & 341 \\
\midrule
\multicolumn{5}{l}{\emph{Optimization}} \\
Train batch size                      & 64       & 64       & 64        & 32 \\
Group size $G$ (RLVR / RLDS)          & 10 / 6   & 10 / 6   & 10 / 6    & 8 / 6 \\
Max episode turns                     & 10       & 5        & 30        & 10 \\
Max assistant / total (tokens)        & 1k / 4k  & 1k / 4k  & 8k / 8k   & 2k / 16k \\
Sampling temperature                  & 0.7      & 0.7      & 1.0       & 1.0 \\
Train epochs                          & 1        & 1        & 20        & 2 \\
\midrule
\multicolumn{5}{l}{\emph{Compute}} \\
Total update steps                    & 312      & 312      & 400       & 315 \\
Wall-clock / step (RLVR / RLDS, s)    & 49.8 / 119.0 & 69.5 / 157.0 & 294.4 / 262.2 & 163.4 / 267.3 \\
Total wall-clock (RLVR / RLDS, h)     & 4.3 / 10.3 & 6.0 / 13.6 & 32.7 / 29.1 & 12.9 / 23.4 \\
Hardware                              & \multicolumn{4}{c}{$8{\times}$H100 80\,GB} \\
\midrule
\multicolumn{5}{l}{\emph{External APIs (RLDS run)}} \\
Tool calls / rollout (mean)           & 0 (no tools) & ${\sim}5$ (local) & 0 (no tools) & ${\sim}3.9$ \\
Total tool calls / run                & 0 & ${\sim}300$k (local FAISS) & 0 & ${\sim}260$k (Serper) \\
Total judge calls / run               & 0 & 0 & 0 & ${\sim}50$--$100$k (\texttt{gpt-5.4-nano}) \\
\bottomrule
\end{tabular}
\end{table}

ScienceWorld's train split is small (1{,}796 examples), so the run is configured for 20 epochs to give the policy enough update steps on the long-horizon rollouts; the high parametric-variation count per task (every example draws a fresh JVM seed) keeps later epochs from being pure replay. FrozenLake and HotpotQA see a single pass over 20k examples at batch 64 ($= 312$ steps); DeepResearch runs 2 epochs over 6{,}480 examples at batch 32 and stops at the 315-step wall-clock cap. The compute-profile reported in Table~\ref{tab:compute-profile} uses these same MLflow numbers; the wall-clock totals here are step time multiplied by step count and exclude data-prep, validation-only steps, and warmup.

\paragraph{R3L / RLDS knobs.} For the reflection-augmented conditions (\textsc{base reflection}, \textsc{rlds}) we use the same R3L recipe across tasks (\texttt{rlef/config/rlef\_grpo.yaml}). Reflection is triggered on the bottom $50\%$ of failures per group; the policy itself acts as both grader and reflector (no external judge model), grading on a 1--5 Likert with an outcome-aware \texttt{pivot\_step}. Auxiliary SFT is enabled with reflection coefficient $0.05$ and retry-distillation coefficient $0.10$, capped at 64 reflect pairs and 64 retry pairs per step. RLDS additionally enables the per-subtask pivot path with a Gaussian kernel ($\sigma{=}1.0$ turn, $\ell^{1}$-normalized per row); the legacy OPMD baseline and positive amplification are disabled.

\section{Subtask Decompositions}
\label{app:taxonomies}

This section lists the subtask taxonomy used by the grader for each task: subtask name, the description shown to the grading model verbatim, and the static importance weight $\varphi_k$ used by SDAE. Weights sum to $1.0$ per task. Names are exactly the strings the grader returns in its JSON output and the strings under which we log per-subtask metrics. The taxonomies live in \texttt{rlef/tasks/<task>/constants.py} as \texttt{SUBTASKS}.

\subsection{FrozenLake}

Five subtasks over a 10-turn budget. Subtasks split spatial reasoning (\texttt{path\_planning}, \texttt{boundary\_awareness}) from hazard reasoning (\texttt{hazard\_identification}, \texttt{dead\_end\_detection}) and from execution-state tracking, so the grader can pin the decisive turn separately for each.

\begin{table}[h]
\centering
\small
\setlength{\tabcolsep}{6pt}
\caption{FrozenLake subtask taxonomy.}
\begin{tabular}{lp{0.62\textwidth}r}
\toprule
Subtask & Description (shown to grader) & $\varphi_k$ \\
\midrule
\texttt{path\_planning}          & Finding a safe route to the goal. & 0.35 \\
\texttt{hazard\_identification}  & Recognizing which tiles are holes. & 0.25 \\
\texttt{state\_position\_tracking} & Maintaining accurate knowledge of current position across steps. & 0.15 \\
\texttt{dead\_end\_detection}    & Identifying configurations where the chosen path will lead to no escape. & 0.15 \\
\texttt{boundary\_awareness}     & Avoiding invalid moves at grid edges. & 0.10 \\
\bottomrule
\end{tabular}
\end{table}

\subsection{HotpotQA}

Four subtasks over a 5-turn budget, mapped to the canonical multi-hop pipeline (\textit{plan} $\to$ \textit{retrieve} $\to$ \textit{extract} $\to$ \textit{ground}). The temporal ordering of pivot steps in Figure~\ref{fig:subtasks-hotpotqa} matches this pipeline: \texttt{search\_planning} pins early, \texttt{answer\_grounding} pins late.

\begin{table}[h]
\centering
\small
\setlength{\tabcolsep}{6pt}
\caption{HotpotQA subtask taxonomy.}
\begin{tabular}{lp{0.62\textwidth}r}
\toprule
Subtask & Description (shown to grader) & $\varphi_k$ \\
\midrule
\texttt{search\_planning}    & Decomposing the multi-hop question and choosing what to look up first; turning each sub-question into a focused initial query rather than passing the question verbatim. Load-bearing in the first tool-call turn. & 0.25 \\
\texttt{evidence\_extraction} & Pulling the right facts out of retrieved passages; identifying which passage actually contains the bridge entity or answer span and not over-trusting the first plausible result. & 0.30 \\
\texttt{search\_iteration}    & Adapting the search plan in response to what came back; feeding bridge entities into the next query, broadening when too little is returned, narrowing when too much, knowing when to stop. & 0.30 \\
\texttt{answer\_grounding}    & Tying the boxed answer to specific retrieved evidence; not falling back on prior knowledge when retrieval came back empty or off-topic. & 0.15 \\
\bottomrule
\end{tabular}
\end{table}

\subsection{ScienceWorld}

Six subtasks over a 30-turn budget --- the longest horizon in the suite and the task with the widest spread of pivot-step means. Subtasks split early-trajectory planning (\texttt{goal\_decomposition}, \texttt{exploration\_efficiency}) from mid-trajectory enactment (\texttt{object\_identification}, \texttt{action\_sequencing}) from late-trajectory verification (\texttt{state\_tracking}, \texttt{scientific\_reasoning}).

\begin{table}[h]
\centering
\small
\setlength{\tabcolsep}{6pt}
\caption{ScienceWorld subtask taxonomy.}
\begin{tabular}{lp{0.62\textwidth}r}
\toprule
Subtask & Description (shown to grader) & $\varphi_k$ \\
\midrule
\texttt{goal\_decomposition}    & Breaking the experiment into sequential subgoals. & 0.20 \\
\texttt{object\_identification} & Identifying which objects are needed and where to find them. & 0.20 \\
\texttt{action\_sequencing}     & Performing actions in the correct causal order. & 0.20 \\
\texttt{state\_tracking}        & Monitoring changes in object properties and environment state. & 0.15 \\
\texttt{scientific\_reasoning}  & Applying correct scientific principles to the experiment. & 0.15 \\
\texttt{exploration\_efficiency} & Navigating and exploring the environment without wasted steps. & 0.10 \\
\bottomrule
\end{tabular}
\end{table}

\subsection{DeepResearch}

Five subtasks over a 10-turn budget. Subtasks separate the search loop (\texttt{search\_planning}, \texttt{search\_iteration}) from the source-handling loop (\texttt{source\_selection}, \texttt{claim\_grounding}) and isolate \texttt{synthesis} as the cross-cutting completeness signal. The temporal stages emerge in Figure~\ref{fig:subtasks-deepresearch}.

\begin{table}[h]
\centering
\small
\setlength{\tabcolsep}{6pt}
\caption{DeepResearch subtask taxonomy.}
\begin{tabular}{lp{0.62\textwidth}r}
\toprule
Subtask & Description (shown to grader) & $\varphi_k$ \\
\midrule
\texttt{search\_planning}  & Issuing focused initial queries that match each sub-question to literature vs.\ open-web evidence. & 0.15 \\
\texttt{source\_selection} & Reading the most authoritative on-topic sources and skipping duplicative or off-topic results. & 0.20 \\
\texttt{search\_iteration} & Reformulating queries when results miss the question and broadening or narrowing scope based on what came back. & 0.30 \\
\texttt{claim\_grounding}  & Wrapping every substantive claim in \texttt{<cite id="..."{>}} whose snippet actually supports the wrapped text. & 0.15 \\
\texttt{synthesis}         & Covering every sub-question and triangulating across sources rather than stitching quotes from one paper. & 0.20 \\
\bottomrule
\end{tabular}
\end{table}

\section{Benchmark Details}
\label{app:benchmarks}

Per-task action spaces, observation formats, reward functions, and termination conditions referenced in \S\ref{sec:exp}. FrozenLake and ScienceWorld are described in full below; HotpotQA and DeepResearch are summarized briefly as they follow established setups.

\subsection{FrozenLake}

\paragraph{Action space.} Free-form text. Each turn the agent emits a single \texttt{<action>D</action>} tag where \texttt{D} is one of \texttt{Up}, \texttt{Down}, \texttt{Left}, \texttt{Right}. There is no tool-calling layer; the interaction parses the tag and executes the corresponding grid move. Actions outside the four cardinal directions, missing tags, and malformed responses are recorded as invalid and consume the turn without changing state. Boundary moves leave the agent in place.

\paragraph{Observation.} The initial observation contains the rendered grid using abstract symbols (\texttt{A}=agent, \texttt{B}=goal, \texttt{C}=hole, \texttt{D}=safe tile) together with a goal description. Symbols are deliberately abstract --- without semantic priors --- so the model must learn the mapping from interaction. Subsequent turns return the updated grid, contextual notes (boundary hits, invalid actions), and the remaining turn budget. Terminal observations name the cause of termination (\textit{reached the goal}, \textit{fell into the hole}, \textit{hit the max step limit}).

\paragraph{Reward.} Sparse binary outcome: $+1$ on reaching the goal tile, $0$ otherwise. There is no shaping, step cost, or partial credit. Goal detection is read from the interaction's authoritative \texttt{turn\_scores} signal; text-based detection is used only as a last-resort offline fallback because the model can otherwise game the reward by emitting the terminal phrase.

\paragraph{Termination.} An episode ends as soon as any of: (i) the agent steps on the goal, (ii) the agent steps on a hole, (iii) the turn budget (10) is exhausted, or (iv) the response token budget is insufficient for the next observation.

\paragraph{Dataset.} Procedurally generated. Maps vary in size between $5{\times}5$ and $8{\times}8$ and in frozen-tile probability $p \in [0.65, 0.85]$; each generated map is validated by depth-first search to admit a path from start to goal of length $\leq 12$. The default split is $20{,}000$ training maps and $500$ validation maps. Random-board generation is the default, but the dataset format also supports explicit board serialization for fixed-eval scenarios.

\paragraph{Runtime.} \texttt{max\_user\_turns} = 10, sampling temperature = 0.7, max prompt length = 1024 tokens, max response length = 4096 tokens.

\subsection{HotpotQA}

\paragraph{Action space.} OpenAI-style function calling in Hermes format. A single tool is exposed:
\begin{itemize}
  \item \texttt{search(query: str, top\_k?: int $\in [1,50]$)} --- queries an E5 dense retriever backed by a FAISS index over the Search-R1 / ERL Wikipedia corpus. Default \texttt{top\_k}~$=5$.
\end{itemize}
The terminal action is to emit the final answer in \verb|\boxed{...}| anywhere in a free-form response. Tool calls and the boxed-answer terminal are mutually exclusive per turn.

\paragraph{Observation.} Hermes-formatted tool returns enumerate retrieved passages as \texttt{[1] <title> <snippet>}, \texttt{[2] ...}, with a truncation notice when results exceed the per-result cap (1{,}200 chars) or the per-call total cap (8{,}000 chars). Non-tool, non-terminal observations are concise prompts to either continue searching or submit an answer. Terminal observations report the outcome explicitly: \textit{``Your answer is correct''} (EM~$=1$), \textit{``partially correct (F1~score: $x$)''} (F1~${\geq}0.3$), or \textit{``incorrect''}.

\paragraph{Reward.} Computed against the normalized gold answer (lowercase, articles and punctuation stripped) at the boxed-answer turn:
\begin{itemize}
  \item Exact-match reward $1.0$ when the normalized prediction equals the normalized gold.
  \item Otherwise token-level F1 against the gold; if F1~${\geq}0.3$ the reward equals F1, else reward is $0$.
\end{itemize}
Episodes that never emit a boxed answer receive $0$. There are no process or penalty terms; tool calls are free.

\paragraph{Termination.} Episodes end when (i) the agent submits a boxed answer, (ii) the 5-turn budget is exhausted, or (iii) the response token budget cannot fit the next observation (estimated at 3 chars/token plus a 20-token overhead).

\paragraph{Dataset.} The HotpotQA distractor split from Hugging Face (\texttt{hotpotqa/hotpot\_qa}). The default training pool is ${\sim}90$\,k examples; we evaluate on a fixed $500$-example validation slice. The retrieval setup (E5 encoder, FAISS index over the Search-R1 Wikipedia dump) follows Search-R1~\citep{jin2025searchr1}.

\paragraph{Runtime.} \texttt{max\_user\_turns}~$=5$, sampling temperature~$=0.7$, max prompt length~$=1{,}024$ tokens, max response length~$=4{,}096$ tokens.

\subsection{ScienceWorld}

\paragraph{Action space.} Free-form text wrapped in \texttt{<action>...</action>} tags, dispatched to the ScienceWorld JVM via py4j. The simulator accepts 18 templated commands: \texttt{open}/\texttt{close OBJ}, \texttt{activate}/\texttt{deactivate OBJ}, \texttt{connect OBJ to OBJ}, \texttt{disconnect OBJ}, \texttt{use OBJ [on OBJ]}, \texttt{look around}, \texttt{examine OBJ}, \texttt{look at OBJ}, \texttt{read OBJ}, \texttt{move OBJ to OBJ}, \texttt{pick up OBJ}, \texttt{pour OBJ into OBJ}, \texttt{mix OBJ}, \texttt{teleport to LOCATION}, \texttt{focus on OBJ}, and \texttt{wait}. Two meta-commands --- \texttt{help} and \texttt{objects} --- are free and return the simulator's enumeration of currently valid actions and referenceable entities respectively. Unrecognized commands return rejection markers (\textit{No known action matches}, \textit{Unknown action}, \textit{Ambiguous request}) and the interaction surfaces matching valid verbs to aid self-correction.

\paragraph{Observation.} Natural-language descriptions emitted by the JVM, covering the current location, inventory, and immediately observable objects. The first observation concatenates the task instruction and reset state. Each subsequent observation is appended with the standard formatting reminder (\texttt{<think>...</think><action>...</action>}). Terminal observations include a steps-used count, the final score on a 0--100 scale, per-step reward attribution, and a qualitative outcome label.

\paragraph{Reward.} The base signal is the simulator's monotonic-max partial-credit score, normalized to $[0,1]$. Two multiplicative discounts and two small additive shaping terms are applied:
\begin{itemize}
  \item \emph{Stagnant-score discount} ($\times 0.5$): triggered when the score reaches a positive value but never improves thereafter and ends below 1.0. Targets binary-guess and prior-knowledge exploitation where the agent collects an early sub-goal and then idles.
  \item \emph{No-experiment discount} ($\times 0.25$): triggered when a procedure-required task scores ${\geq}0.5$ but the agent never executes any experimental verb (\texttt{connect}, \texttt{activate}, \texttt{deactivate}, \texttt{use}, \texttt{mix}, \texttt{pour}, \texttt{wait}). Mitigates the \textit{focus-on-X then move-to-box} reward hack on tasks like \texttt{test-conductivity} where the simulator's partial-credit scorer awards classification sub-goals even when the required experiment was not performed. Classification task families (\texttt{find-*}, \texttt{lifespan-*}, \texttt{identify-life-stages}) are exempt because \texttt{focus} is the terminal action by design.
  \item \emph{Reasoning-absence penalty} ($-0.1$ when \texttt{<think>} tags appear in fewer than half of the turns) and \emph{action-diversity bonus} (up to $+0.05$ for varied action verbs). These are reusable components shared across tasks.
\end{itemize}

\paragraph{Termination.} An episode ends when the JVM signals task completion, the turn budget (30) is exhausted, the same action is emitted three turns in a row (early-termination guard against degenerate loops), or an internal JVM error occurs (the episode returns score 0.0 gracefully).

\paragraph{Dataset.} ScienceWorld~\citep{wang2022scienceworld} provides 30 procedural tasks across 10 science topics with parametric variations. We use simplification mode \texttt{easy}. We support two splits: (i) the per-task train/dev split distributed with the benchmark, and (ii) an R3L-style task-generalization split with 17 train tasks and 13 held-out test tasks. Per-task variations are sampled at 50\% by default.

\paragraph{Runtime.} \texttt{max\_user\_turns} = 30 (matching the JVM step limit), sampling temperature = 1.0, max prompt length = 8192 tokens, max response length = 8192 tokens.

\subsection{DeepResearch}

\paragraph{Action space.} OpenAI-style function calling in Hermes format with four tools, all backed by the Serper.dev infrastructure plus local PDF/web fetchers:
\begin{itemize}
  \item \texttt{search\_papers(query: str, limit?: int $\in [1,10]$)} --- Google Scholar via Serper. Internally Serper is always called with $\texttt{max}{=}10$ to maximize cache hits and the model-visible result list is truncated to \texttt{limit} (default $5$).
  \item \texttt{read\_paper(pdf\_url: str)} --- downloads the PDF locally with a browser-like User-Agent, validates content type and magic bytes (paywall detection), and converts to markdown. Hard cap $15$\,MB.
  \item \texttt{google\_search(query: str, gl?: str, hl?: str)} --- open-web search via Serper, defaulting to 10 organic results.
  \item \texttt{browse\_webpage(url: str)} --- Serper scrape returning markdown for a single URL. Default per-page char cap $20{,}000$.
\end{itemize}
All tools share a $120$\,s wall-clock cap and exponential-backoff retries. Outputs are cached in a Ray actor (LRU, max $16{,}384$ entries) shared across rollout workers. The Serper tools share a per-process semaphore (default $12$/worker, ${\sim}96$ global); \texttt{read\_paper} runs locally with a separate, larger semaphore. The terminal action is closing an \verb|<answer>...</answer>| block; substantive claims inside the answer must be wrapped in \verb|<cite id="...">| with ids returned by retrieval tools.

\paragraph{Observation.} Tool results are emitted as \texttt{[id=<id>] <title> ... [cites=N]} headers followed by snippet/content; paper ids use the prefix \texttt{p\_<sha1>} and web ids are URL hashes. Truncation notices are appended when char caps fire. Terminal observations report the rubric score on a $[0,1]$ scale plus verdict-bucket counts (\textit{``$F$ fully met, $P$ partial, $Z$ not met''}) over the $N$ rubric criteria. Criterion titles and grader hints are deliberately \emph{omitted} from the terminal observation to prevent answer leakage into reflection / aux-SFT.

\paragraph{Reward.} A weighted sum of one outcome and three auxiliary signals, each in $[0,1]$:
\begin{align*}
r &= 0.7\,r_{\text{rubric}} + 0.1\,r_{\text{citation}} + 0.1\,r_{\text{format}} + 0.1\,r_{\text{search}}, \\
r_{\text{citation}} &= 0.6\,F_1^{\text{cite}} + 0.4\,r_{\text{citation-format}}.
\end{align*}
\begin{itemize}
  \item $r_{\text{rubric}}$: an LLM judge (default \texttt{gpt-5.4-nano} with low reasoning effort) grades the final answer against the per-instance static rubric on a 0--5 scale per criterion, normalized by the sum of criterion weights.
  \item $F_1^{\text{cite}}$: per-claim F1 between each \verb|<cite id>|'d span and the cited snippet, evaluated by an LLM judge and averaged. Toggleable via env var (\texttt{DEEPRESEARCH\_CITATION\_F1\_ENABLED}); when off, $r_{\text{citation}}$ falls back to $r_{\text{citation-format}}$ alone.
  \item $r_{\text{citation-format}}$: fraction of cited ids that were actually retrieved by a tool call (anti-hallucination check on ids).
  \item $r_{\text{format}} = 0.5 \cdot \mathbf{1}[\text{has answer block}] + 0.3 \cdot \mathbf{1}[\text{has cite}] + 0.2 \cdot \mathbf{1}[\text{has tool call}]$.
  \item $r_{\text{search}} = \min(n_{\text{tool calls}} / 3, 1)$.
\end{itemize}
A length penalty (soft target $6{,}000$ chars, hard cap $15{,}000$, max penalty configurable) is available but disabled by default.

\paragraph{Termination.} Episodes end when (i) the agent closes an \verb|<answer>| block, (ii) the response token budget cannot fit the next feedback (3 chars/token estimate, 20-token overhead), (iii) the prompt buffer fills, or (iv) the 10-turn budget is exhausted. All terminal paths invoke the rubric judge subject to a process-wide $90$\,s wall-clock cap.

\paragraph{Dataset.} A merged pool of five Hugging Face sources, all distributed with static per-instance rubrics: \texttt{rl-research/dr-tulu-rl-data} (OpenScholar, AstaBench-ScholarQA-CS2, SearchArena slices, capped at $2$\,k / $2$\,k / $5$\,k), \texttt{anisha2102/RaR-Science-20k-o3-mini} ($1.5$\,k), and \texttt{anisha2102/RaR-Medicine-20k-o3-mini} ($0.5$\,k). Heterogeneous rubric schemas are normalized to \{\texttt{id}, \texttt{title}, \texttt{description}, \texttt{weight}, \texttt{leak\_safe\_hint}\}; leak-safe hints are LLM-generated to support auxiliary distillation without exposing the gold answer. Setup otherwise follows DR Tulu~\citep{shao2025drtulu}; the evolving-rubric mechanism from the original paper is intentionally omitted (rubrics are static).

\paragraph{Runtime.} \texttt{max\_user\_turns}~$=10$, sampling temperature~$=1.0$, max prompt length~$=2{,}048$ tokens, max response length~$=16{,}384$ tokens.

\section{Reflection and Grading Prompts}
\label{app:prompts}

This section reproduces the prompts used by $\pi_\theta$ in the reflect--retry loop on HotpotQA. The same template structure (system prompt, rollout-time conditioning, R3L two-pass grading, R3L graded reflection) is used on every task; only the rubric strings (Appendix~\ref{app:taxonomies}) and the system prompt are task-specific. All prompts live verbatim in \texttt{rlef/tasks/<task>/constants.py} and substitution placeholders (e.g.\ \texttt{\{trajectory\_summary\}}, \texttt{\{sub\_tasks\_section\}}) are filled by \texttt{rlef/r3l/credit.py} at rollout time.

\paragraph{System prompt (HotpotQA).}
\begin{small}
\begin{verbatim}
You are a helpful assistant who answers questions directly and efficiently.
When you have enough evidence, provide your final answer inside \boxed{}.

Use the search tool when you need evidence.
You may optionally pass top_k to the search tool when you need broader retrieval,
but prefer smaller top_k values unless more coverage is necessary because long
search responses may be truncated.
When calling tools, pass valid JSON arguments with double-quoted keys and values,
for example: {"query":"who wrote the iliad","top_k":3}.
Do not output tool schemas, pseudo-JSON, or commentary in tool arguments.
\end{verbatim}
\end{small}

\paragraph{Grading prompt (R3L two-pass, all tasks).} The grader sees the full trajectory plus the per-task subtask list (rendered from \texttt{SUBTASKS} as a numbered \texttt{\{sub\_tasks\_section\}}) and emits one Likert grade and one outcome-aware \texttt{pivot\_step} per subtask. The pivot semantics are explicitly outcome-aware: on a successful trajectory the pivot is the earliest \emph{load-bearing} step for the subtask; on a failure it is the earliest \emph{degradation} step.

\begin{small}
\begin{verbatim}
You are an evaluator that grades an agent's trajectory across competency
dimensions and identifies, for each, the single step where that competency
was most *pivotal* for the outcome that occurred. Score each dimension on a
1-5 scale where 1=critical failure, 2=poor, 3=adequate, 4=good, 5=excellent.

A "pivot_step" is outcome-aware:
- If the trajectory succeeded ('success' or 'success_but_inefficient'),
  the pivot_step is the earliest step where the agent's handling of this
  competency was load-bearing for success - a decision that, if reversed
  or substituted with a plausible alternative, would have meaningfully
  reduced the chance of reaching the correct answer.
- If the trajectory failed ('failure'), the pivot_step is the earliest
  step where this competency first degraded - the point at which the
  root cause of the failure manifested along this dimension.

Use null when no specific step is pivotal for this competency.

{sub_tasks_section}

## Trajectory
{trajectory_summary}

Please output ONLY a single valid JSON object in the following format:
{
  "trajectory_outcome": "One of: 'success', 'success_but_inefficient', 'failure'",
  "root_cause_analysis": "If the trajectory failed, explain why. If it
    succeeded, explain what made it succeed. Trace observable symptoms back
    to a fundamental cause using iterative 'why' questioning.",
  "sub_task_grades": [
    {
      "sub_task": "Name of the competency dimension from the list above.",
      "score": "Integer 1-5 (1=critical failure, 5=excellent).",
      "assessment": "One-sentence justification for the score.",
      "pivot_step": "Integer step index where this competency was pivotal,
        with outcome-aware semantics described above, or null."
    }
  ]
}

Include one entry in sub_task_grades for EVERY competency dimension above.
\end{verbatim}
\end{small}

\paragraph{Graded-reflection prompt (R3L two-pass, all tasks).} After grading, the same policy is asked to write a reflection conditioned on the grades. The output is a structured JSON object with a \texttt{retry\_from\_step} field that the trainer feeds into the per-subtask pivot mask.

\begin{small}
\begin{verbatim}
You are a Reflector analyzing a trajectory that has already been graded
across competency dimensions. Use the grades to inform your analysis.

{grading_section}

{memory_section}{contrastive_section}
## Failed trajectory
{trajectory_summary}

Analyze the trajectory in light of the grades above. Identify which
subtask(s) to address and produce a reflection that will most improve
performance on a retry.

Please output ONLY a single valid JSON object in the following format:
{
  "trajectory_summary": "Concise overview covering: (1) strategy employed,
    (2) final result, (3) key observations about execution quality.",
  "root_cause_analysis": "Deep causal analysis grounded in specific
    trajectory steps. Walk forward and name the earliest step where the
    agent had a real choice and picked wrong; quote that step's action
    or observation. Chain reasoning explicitly from that decisive moment
    to the final outcome.",
  "trajectory_outcome": "One of: 'success', 'success_but_inefficient',
    'failure'",
  "improvement_suggestion": "Actionable principle for the retry attempt.",
  "retry_from_step": "Non-negative integer naming the earliest step where
    a different action would have materially changed the outcome. Quote
    that step's action or observation in root_cause_analysis. Use 0 when
    the first action itself was the decisive wrong choice."
}
\end{verbatim}
\end{small}

The corresponding ScienceWorld, FrozenLake, and DeepResearch prompts share this structure verbatim; the only differences are the system prompt text and the contents of \texttt{\{sub\_tasks\_section\}} (Appendix~\ref{app:taxonomies}). The single-pass variant \texttt{R3L\_REFLECTION\_PROMPT} (used when the grading pass is disabled) drops the \texttt{\{grading\_section\}} block but preserves the same JSON output schema sans \texttt{sub\_task\_grades}.

\begin{figure}[h]
\centering
\begin{small}
\begin{minipage}{0.95\textwidth}
\textbf{System prompt (excerpt).}
\begin{verbatim}
You are an agent playing a game on a grid, acting as a reasoning engine.

# Game Rules
*(To be discovered through play - consult any additions below.)*

# Strategic Playbook
*(To be discovered through play - consult any additions below.)*

# Required response structure
<think>
**1. Rules & Playbook Consulted:** ...
**2. Analysis of the Current State:** ...
**3. Prediction of the Value of Current States:** ...
**4. Prediction of Immediate Consequences:** ...
</think>
<action>Up | Down | Left | Right</action>
\end{verbatim}

\textbf{User turn 1 (initial observation).}
\begin{verbatim}
Current Observation:
D D D D D D C
D D C D D D D
D D B D D C D
C D D C D D D
D C C C C A C
D D D C D C C
C C D C D D C
You have not achieved the goal yet. Please give the next action.
\end{verbatim}

\textbf{Assistant turn 1 (response, abridged).}
\begin{verbatim}
**1. Rules & Playbook Consulted:** None available. The presence of 'C',
'D', 'B', and 'A' suggests possible progression rules ... however,
without explicit game rules, the interpretation is speculative.

**4. Prediction of Immediate Consequences:**
1. Action: Right - leads to (3,3) ('D'), opens path toward center.
2. Action: Down - leads to (4,2) ('C'). 'C' is adjacent to 'A' in
   row 5, column 5. A path C->C->C->C->A suggests a direct route to 'A'.

Both actions are plausible, but down leads more directly toward a
contiguous path to 'A' via 'C' tiles.

<action>Down</action>
\end{verbatim}

\textbf{User turn 2 (terminal observation).}
\begin{verbatim}
The agent moved onto a 'C' tile and fell into the hole.
\end{verbatim}
\end{minipage}
\end{small}
\caption{Anecdotal FrozenLake rollout at eval step 30. The empty \texttt{Game Rules} block leaves the policy without the \texttt{C}-is-a-hole prior, and the abstract-symbol vocabulary (\S\ref{app:benchmarks}) prevents pretraining from filling the gap. The grader pins \texttt{hazard\_identification} (highest weight after \texttt{path\_planning}) as the failed competency and the reflector emits an addition that fixes the rule for the retry.}
\label{fig:anecdote}
\end{figure}

\end{document}